\documentclass[10pt,twocolumn]{article}

\usepackage[utf8]{inputenc}
\usepackage[T1]{fontenc}
\usepackage[margin=17.5mm]{geometry}
\usepackage{amsmath,amssymb,amsfonts,amsthm}
\usepackage{graphicx}
\usepackage{booktabs}
\usepackage{cite}
\usepackage{hyperref}
\usepackage{microtype}
\usepackage{xcolor}
\usepackage{url}
\usepackage{algorithm}
\usepackage{algpseudocode}
\usepackage{multirow}
\usepackage{placeins}
\hypersetup{
    colorlinks=true,
    linkcolor=blue!70!black,
    citecolor=blue!70!black,
    urlcolor=blue!70!black
}

\newtheorem{proposition}{Proposition}

\title{\textbf{\Large Universal Procedural Boundary Decision Map: Constant-Storage Real-Time Edge Triage Across Heterogeneous Domains}\thanks{Version~3.0 standardized preprint (superseding Zenodo v2.0 \texttt{10.5281/zenodo.22939253} and \texttt{arXiv:2609.25498v2}), synchronized with FNS v4.0 (\texttt{arXiv:2611.32393}) and OED v4.0 (\texttt{arXiv:2609.30115}).}}

\author{
    \textbf{Volkan Da\u{g}l\i}\textsuperscript{1,2,*} \quad \textbf{Zerrin Da\u{g}l\i}\textsuperscript{3,4} \quad \textbf{Da\u{g}han Da\u{g}l\i}\textsuperscript{5} \\[0.12cm]
    \footnotesize\textsuperscript{1}\textit{Anadolu University, Eski\c{s}ehir, T\"{u}rkiye} $\;\bullet\;$ \textsuperscript{2}\textit{ITouch Systems, \c{C}ukurova University Technopark, Adana, T\"{u}rkiye} \\
    \footnotesize\textsuperscript{3}\textit{Mersin University, Mersin, T\"{u}rkiye} $\;\bullet\;$ \textsuperscript{4}\textit{Yusuf Kalkavan Anatolian High School, Ministry of National Education, Mersin, T\"{u}rkiye} \\
    \footnotesize\textsuperscript{5}\textit{Toros Science High School, MEV (Toros University), Mersin, T\"{u}rkiye} \\
    \footnotesize\textsuperscript{1,2}\textit{ORCID: 0009-0000-1587-8703} \quad
    \textsuperscript{3,4}\textit{ORCID: 0000-0001-9490-6425} \quad
    \textsuperscript{5}\textit{ORCID: 0009-0003-2492-8313} \\
    \footnotesize\textsuperscript{*}\textit{Corresponding author: Volkan Da\u{g}l\i} \quad \textit{Code:} \url{https://github.com/pCwOrM/werr} \\
    \footnotesize\textit{Open Dataset:} \url{https://huggingface.co/datasets/pCwOrM/werr_open_decisions}
}

\date{November 30, 2026}

\begin{document}

\maketitle

\begin{abstract}
Real-time operational triage---API gateway gating, fraud filtering, robotics safety interlocks, and smart-contract circuit breakers---requires deterministic, strongly typed sub-10\,ms verdicts, yet transformer language models incur multi-gigabyte VRAM footprints, stochastic drift, and heavy memory-movement overhead. We present \textbf{WERR Map v3.0}, a constant-storage procedural System-1 triage engine operationalizing Fractal Neural Synthesis~\cite{dagli2026fns} and Orbital Error Dynamics~\cite{dagli2026oed} by replacing persistent neural weight tensors with a $24\text{-byte}$ seed $\Theta = (c_x, c_y, \mathrm{zoom}) \in \mathbb{F}_{64}^3$ evaluated on the complex quadratic boundary $\partial\mathcal{M}$ over a Quadro ($2{\times}2$) aperture and a cyclic $\mathbb{Z}/9\mathbb{Z}$ ($36{\times}36$) grid. WERR combines bilingual state-to-phase projection ($\Phi_D$) and MD5 instruction phase shift with weighted 3-scale Harmonic Tripod fusion, multiplicative streaming EMA calibration, and coupled pitchfork margin expansion, while a semantic damping pre-filter ($\mathcal{T}_{\mathrm{desc}} = 0.045$) cuts escape depth by $45.8\%$ ($2.5\times$ speedup; $0/10$ decoy-injection probes bypassed, $95\%$ exact CI $[0.0\%, 30.9\%]$). Across six SHA-256 sealed artifacts at $0\text{ B}$ VRAM, WERR achieves $411.9\text{ moves/s}$ on Snake~AI ($1.32\text{ ms}$ p50; $10.2\times$--$492.8\times$ lower latency than ModernBERT and cloud LLMs), $49/49$ ($100\%$) on WindTunnel WebMCP ($1.81\text{ ms}$ p50), $38.0\times$ speedup on Jevenator~2 ($0$ false positives), and $100\%$ TypeSafe wire compliance on air-gapped JevBench v1.4.1 ($N=231$; $53.25\%$ sealed Tripod, $54.55\%$ repo Universal Cusp, $85.42\%$ Easy) with zero task-specific rules. In repository ablations, a $120\text{ B}$ five-domain atlas reaches $93.1\%$ macro / $92.6\%$ micro accuracy ($N=326$, $z=8.92, p < 10^{-18}$). We further audit \texttt{pCwOrM/werr\_open\_decisions} on Hugging Face ($4{,}369$ records, $10{,}968$ typed queries across $67$ active days; pure System-1 p50 $= 8.64\text{ ms}$, $n=2{,}048$).
\end{abstract}

\vspace{0.12cm}

\section{Introduction}
\label{sec:intro}

Autonomous software agents, industrial controllers, and decentralized execution layers increasingly confront a structural mismatch between \emph{computational capacity} and \emph{operational cadence}. Following Kahneman's dual-process distinction \cite{kahneman2011thinking}, modern Large Language Models (LLMs) excel at deliberative System-2 reasoning---multi-step synthesis, open-ended drafting, and nuanced planning---yet are routinely invoked for high-frequency System-1 triage: deciding whether an incoming API payload is malicious, selecting one of three tool routes, grading anomaly severity on an ordinal scale, or tripping a latency-critical safety interlock.

Invoking dense transformer architectures \cite{vaswani2017attention, touvron2023llama} on sub-10\,ms System-1 control loops introduces four acute operational bottlenecks:
\begin{enumerate}
    \item \textbf{Memory-Movement Bottleneck:} Modern neural weights occupy $O(W)$ persistent storage ($10^8$--$10^{11}$ parameters). Fetching weights from off-chip DRAM consumes two to three orders of magnitude more energy than on-chip arithmetic \cite{horowitz2014computing}, creating substantial energy and latency overhead for routine routing gates \cite{strubell2019energy}.
    \item \textbf{Latency Variance and Cloud Dependence:} Remote API round-trips span $300$--$1{,}500\text{ ms}$, while compact local $421\text{M}$--$4\text{B}$ models \cite{warner2024modernbert} still demand $944\text{ MB}$--$8\text{ GB}$ of VRAM or unified memory, precluding execution inside strict zero-VRAM containers, microcontrollers, or deterministic smart-contract virtual machines.
    \item \textbf{Non-Deterministic Drift and Schema Fragility:} Autoregressive decoding requires structured-output parsers and temperature control, yet remains susceptible to prompt-injection overrides \cite{perez2022ignore} and schema drift across minor model revisions.
    \item \textbf{Lack of Auditable Open Telemetry:} Proprietary agent guardrails rarely publish reproducible, strongly typed decision logs capturing both sub-10\,ms edge gating and System-2 escalation behavior across multilingual production traffic.
\end{enumerate}

In our companion theoretical works---Fractal Neural Synthesis (FNS v4.0 \cite{dagli2026fns}), Orbital Error Dynamics (OED v4.0 \cite{dagli2026oed}), and machine-checked Lean~4 algebraic verification \cite{dagli2026lean4oed}---we established that complex quadratic boundary dynamics ($\partial\mathcal{M}$) can act as a constant-storage procedural reservoir parameterized by a $24\text{-byte}$ seed $\Theta = (c_x, c_y, \mathrm{zoom}) \in \mathbb{F}_{64}^3$. Here, we present \textbf{WERR Map v3.0} (Fig.~\ref{fig:architecture}), the applied System-1 natural-language and structured-state decision architecture that operationalizes this procedural geometry for real-time edge triage at $0\text{ B}$ VRAM.

\subsection{Contributions and Revision Scope}
\begin{figure*}[t]
\centering
\includegraphics[width=0.98\textwidth]{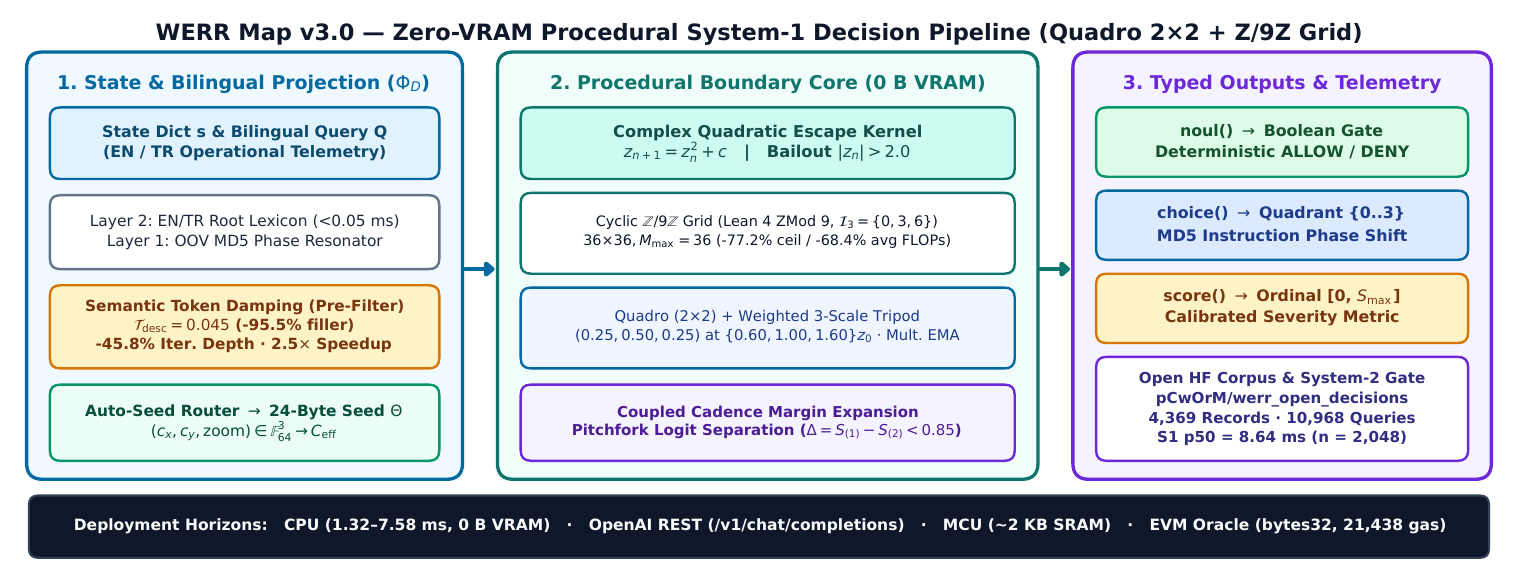}
\caption{WERR Map v3.0 end-to-end procedural System-1 pipeline. Stage~1 maps structured operational state dictionaries and bilingual (EN/TR) queries via a sub-$0.05\text{ ms}$ lexicon, OOV MD5 phase resonator, optional semantic damping pre-filter ($\mathcal{T}_{\mathrm{desc}}=0.045$), and Auto-Seed Router onto a $24\text{-byte}$ boundary seed $\Theta$. Stage~2 evaluates the complex quadratic escape kernel on a cyclic $\mathbb{Z}/9\mathbb{Z}$ ($36{\times}36, M_{\max}=36$) Quadro ($2{\times}2$) grid with weighted 3-scale Harmonic Tripod fusion, multiplicative EMA calibration, and coupled pitchfork logit separation at $0\text{ B}$ VRAM. Stage~3 emits strongly typed \texttt{noul()}, \texttt{choice()}, and \texttt{score()} verdicts and logs optional open telemetry to \texttt{pCwOrM/werr\_open\_decisions}.}
\label{fig:architecture}
\end{figure*}

This work makes five primary contributions while deliberately preserving full backward compatibility with the public Quadro ($2{\times}2$) + $\mathbb{Z}/9\mathbb{Z}$ reference engine in \texttt{pCwOrM/werr}:
\begin{enumerate}
    \item \textbf{Constant-Storage System-1 Decision Architecture:} Formalizes the end-to-end procedural pipeline (Fig.~\ref{fig:architecture}, Algorithm~\ref{alg:werr_step}), uniting bilingual state-to-phase projection ($\Phi_D$), boundary-corrected Quadro ($2{\times}2$) escape density on a $36{\times}36$ cyclic $\mathbb{Z}/9\mathbb{Z}$ grid, weighted three-scale Harmonic Tripod fusion, multiplicative streaming EMA calibration, and coupled pitchfork margin expansion.
    \item \textbf{Decoupled Air-Gapped Reproducibility:} Strictly separates the stateless procedural core (\texttt{werr}, Layer~B, zero network imports, \texttt{WERR\_TELEMETRY=0} by default) from optional serving gateways (\texttt{answerr Pro}, Layer~A), audited via $38/38$ unit tests and $100/100$ bit-exact deterministic replay.
    \item \textbf{Cryptographically Sealed External Validation:} Evaluates WERR across six SHA-256 sealed artifacts spanning Snake~AI real-time control, WindTunnel WebMCP tool routing \cite{windtunnel2026webmcp}, Jevenator~2 visual triage \cite{jevenator2026}, and air-gapped JevBench v1.4.1 \cite{standhartinger2026jevbench}, alongside transparent separation from unsealed repository ablations.
    \item \textbf{Open Bilingual Hugging Face Telemetry Audit:} Audits the live production growth of \texttt{pCwOrM/werr\_open\_decisions} \cite{dagli2026werrdataset} across $4{,}369$ records and $10{,}968$ typed evaluations over $67$ active days, reconciling every record and separating pure System-1 gating ($8.64\text{ ms}$ p50, $n=2{,}048$) from deliberative System-2 orchestration ($45.63\text{ ms}$ p50, $n=2{,}074$).
    \item \textbf{Exact Code-to-Theory Alignment:} Synchronizes every equation, algorithm step, calibration metric, and statistical confidence interval (Wilson \cite{wilson1927probable}, Clopper--Pearson \cite{clopper1934use}, and two-proportion $z$-tests) directly against the shipped \texttt{pCwOrM/werr} implementation.
\end{enumerate}

\paragraph{Revision Note (v2.0 to v3.0).}
Compared to v2.0 (\texttt{10.5281/zenodo.22939253}), this standardized v3.0 release harmonizes mathematical terminology with FNS v4.0 \cite{dagli2026fns} and OED v4.0 \cite{dagli2026oed}, removes unmeasured theoretical energy-ratio extrapolations, expands the audited Hugging Face dataset $3.8\times$ ($1{,}150 \to 4{,}369$ records), reconciles all micro/macro accuracy and calibration metrics against \texttt{SEAL\_MANIFEST.json}, and replaces legacy raster charts with six reproducible vector figures plus an editable \texttt{.drawio} system diagram.

\section{Related Work and Positioning}
\label{sec:related}

WERR Map v3.0 sits at the intersection of three research lineages while occupying a distinct zero-VRAM operational niche:

\paragraph{Indirect Encoding and Procedural Weight Synthesis.}
Compositional Pattern Producing Networks (HyperNEAT \cite{stanley2009hyperneat}) and HyperNetworks \cite{ha2017hypernetworks} generate connection weights from geometric coordinates or auxiliary generator networks, whereas FractalNet \cite{larsson2017fractalnet} uses self-similar macro-architectural wiring while retaining dense stored tensors. Knowledge distillation \cite{hinton2015distilling} and sparsely-gated Mixture-of-Experts \cite{shazeer2017moe} reduce active compute per token yet still require storing hundreds of millions to billions of parameters in memory. By contrast, WERR Map derives transient decision geometry directly from complex quadratic boundary iteration using $24$ persistent bytes per domain coordinate, inheriting the scale-invariant fractal boundary capacity formalized in FNS v4.0 \cite{dagli2026fns} (whose boundary Hausdorff dimension equals $2$ \cite{shishikura1998hausdorff}).

\paragraph{Strongly Typed Agent Protocols and Edge Classifiers.}
Recent TypeSafe AI benchmarks---including JevBench \cite{standhartinger2026jevbench}, WindTunnel WebMCP \cite{windtunnel2026webmcp}, and Jevenator~2 \cite{jevenator2026}---evaluate agent decision engines on strict wire-schema conformance, Expected Calibration Error (ECE) and temperature scaling \cite{guo2017calibration}, latency, and marginal inference cost. Compact encoder classifiers such as ModernBERT-$421\text{M}$ \cite{warner2024modernbert} narrow the latency gap to $\sim 13\text{ ms}$ on unified-memory accelerators, yet still demand $\sim 944\text{ MB}$ of persistent weight memory. WERR Map demonstrates that deterministic boundary geometry can satisfy $100\%$ TypeSafe wire compliance on commodity CPUs at $1.32$--$7.58\text{ ms}$ with $0\text{ B}$ VRAM.

\paragraph{Adversarial Robustness and Deterministic Oracles.}
Prompt-injection attacks exploit the shared token-context window of autoregressive decoders \cite{perez2022ignore}. Because WERR Map lacks a generative prompt-following decoder and attenuates ungrounded descriptive payloads via semantic token damping, imperative text overrides cannot hijack control flow. Furthermore, deterministic fixed-point execution aligns naturally with authenticated smart-contract oracles \cite{zhang2016town} and verifiable off-chain computation \cite{bensasson2014snarks}, enabling single-slot on-chain reflex gating \cite{dagli2026werracle}.

\section{Mathematical Foundations}
\label{sec:math}

\subsection{Constant-Storage Complex Quadratic Kernel}
For a $24\text{-byte}$ coordinate seed $\Theta = (c_x, c_y, \mathrm{zoom}) \in \mathbb{F}_{64}^3$, let $c_0 = c_x + i c_y \in \mathbb{C}$ lie on or adjacent to the boundary $\partial\mathcal{M}$ of the complex quadratic family $f_c(z) = z^2 + c$ initialized at $z_0 = 0$. Given an operational state dictionary $s \in \mathcal{S}$ and bilingual query $Q \in \mathcal{Q}$, a deterministic projection operator $\Phi_D : \mathcal{S} \times \mathcal{Q} \to \mathbb{C}$ embeds $s$ into a bounded 16-dimensional feature vector $\mathbf{v}(s) \in [-1, 1]^{16}$ (via logistic scaling $2/(1+e^{-x/10})-1$ for numeric scalars and MD5 unit-circle phase pairs $(\sin\theta, \cos\theta)$ for OOV strings) plus domain risk polarity, mapping state-query semantics into a bounded perturbation $\delta c = \Phi_D(s, Q)$ around $c_0$:
\begin{equation}
C_{\mathrm{eff}}(s, Q) = c_0 + \frac{\Phi_D(s, Q)}{\mathrm{zoom}}.
\label{eq:ceff}
\end{equation}
Around $C_{\mathrm{eff}}$, WERR evaluates an $N \times N$ spatial grid of complex points $c_{u,v} = C_{\mathrm{eff}} + \Delta_{u,v}(\mathrm{zoom})$ for $u, v \in \{0, \dots, N-1\}$ over $M_{\max}$ iterations with bailout $|z_n| > 2.0$. Defining the integer escape index $K(c_{u,v}) \in \{0, \dots, M_{\max}\}$ (with $K = M_{\max}$ if $|z_k| \le 2.0$ for all $k < M_{\max}$), the normalized escape field implemented in \texttt{werr/fractal.py} is:
\begin{equation}
u(c_{u,v}) = \frac{K(c_{u,v})}{M_{\max}} \in [0, 1].
\label{eq:escape}
\end{equation}

\paragraph{Storage Accounting Clarification.}
WERR stores zero persistent neural weight matrices ($\mathrm{Mem}_{\mathrm{persist}}(\mathbf{W}) = 0\text{ B}$, $0\text{ B}$ GPU VRAM); the parametric boundary representation consists solely of $\mathrm{Mem}_{\mathrm{persist}}(\Theta) = 24\text{ B}$ per seed ($120\text{ B}$ for the five-domain atlas in Table~\ref{tab:atlas}). As in any software runtime, lightweight static preprocessing code, bilingual keyword tables, and state-signature dictionaries ($\sim 75\text{ KB}$ of Python source) reside in standard CPU RAM and are accounted separately from learned neural tensor weights.

\begin{table}[t]
\centering
\caption{Calibrated Five-Domain $24\text{-Byte}$ Coordinate Atlas ($120\text{ B}$ Total; \texttt{werr/gates/})}
\label{tab:atlas}
\resizebox{\columnwidth}{!}{%
\begin{tabular}{@{}lrrr@{}}
\toprule
\textbf{Operational Domain} & \textbf{$c_x$ ($\mathbb{F}_{64}$)} & \textbf{$c_y$ ($\mathbb{F}_{64}$)} & \textbf{$\mathrm{zoom}$} \\
\midrule
API Gateway Security & $-0.74364389$ & $+0.13182590$ & $120.0$ \\
E-Commerce Fraud & $-0.74950000$ & $+0.08200000$ & $70.0$ \\
Financial Risk & $-0.74800000$ & $+0.06500000$ & $60.0$ \\
Game Combat Reflex & $-0.74450000$ & $+0.12500000$ & $65.0$ \\
IoT Life Safety & $-0.74500000$ & $+0.11200000$ & $85.0$ \\
\bottomrule
\end{tabular}}
\end{table}

\subsection{Quadro (\texorpdfstring{$2{\times}2$}{2x2}) Aperture and Strongly Typed Primitives}
To preserve full reproducibility with the public \texttt{pCwOrM/werr} engine, WERR Map v3.0 partitions the $N \times N$ escape field into four equal spatial quadrants---the \textbf{Quadro ($2{\times}2$) Aperture} $\{Q_0, Q_1, Q_2, Q_3\}$ (Fig.~\ref{fig:quadro_ema}a). Because $u(c_{u,v})$ is bounded on $[0, 1]$, naive sample averages suffer boundary truncation bias near $0$ and $1$. Following bounded-domain kernel density correction, \texttt{extract\_bounded\_quadrant\_weights} assigns each pixel an inverse Gaussian-CDF correction weight $W(u, h) = 1 / \operatorname{clip}_{[0.45, 1]}\!\bigl(\Phi(u/h) + \Phi((1{-}u)/h) - 1\bigr)$ with bandwidth $h = 0.12$, and computes each quadrant's composite boundary energy $E_k \in [0, 1]$:
\begin{align}
E_k &= 0.65\,\frac{\sum_{q \in Q_k} W(u_q, h)\,\mathbb{I}[u_q \ge 0.90]}{\sum_{q \in Q_k} W(u_q, h)} \nonumber \\
    &\quad + 0.35\,\frac{\sum_{q \in Q_k} W(u_q, h)\,u_q}{\sum_{q \in Q_k} W(u_q, h)},
\label{eq:quad_energy}
\end{align}
alongside $4{\times}4$ hierarchical quadtree tile ratios for fine-grained feature projections.

All operational outputs are strictly constrained to three strongly typed decision primitives:
\begin{enumerate}
    \item \textbf{Boolean Gate (\texttt{noul}):} Combines state-polarity alignment and boundary-modulated feature projections via logistic gating against threshold $\tau \in (0, 1)$, emitting $\texttt{(bool, } p \in [0.01, 0.99]\texttt{)}$ for deterministic \texttt{ALLOW}/\texttt{DENY} verdicts.
    \item \textbf{Categorical Router (\texttt{choice}):} Scores $m$ candidate action criteria $\{a_1, \dots, a_m\}$ by fusing criteria token/bigram alignment, negation scoping, and quadrant boundary fields, applying temperature-scaled softmax \cite{guo2017calibration} with strict unit-simplex compliance $\left|\sum_{j=1}^{m} p_j - 1\right| < 10^{-3}$ (\texttt{SUM\_TOL}).
    \item \textbf{Ordinal Grader (\texttt{score}):} Projects ordered severity descriptors $\{L_0, \dots, L_{S_{\max}}\}$ onto calibrated ordinal expectations $\hat{y} \in [0, S_{\max}]$ with per-level probability mass.
\end{enumerate}

\subsection{Cyclic \texorpdfstring{$\mathbb{Z}/9\mathbb{Z}$}{Z/9Z} Modular Resonant Grid}
Earlier prototypes (v1.0) evaluated dyadic grids ($N = 64$, $M_{\max} = 50$), incurring a worst-case iteration ceiling of $64^2 \times 50 = 204{,}800$ complex quadratic evaluations per query. In WERR Map v3.0, we adopt $N = 36$ and $M_{\max} = 36$, reducing the worst-case operational ceiling directly through grid and iteration-budget compression:
\begin{equation}
\Delta_{\mathrm{ceiling}} = 1 - \frac{36^2 \times 36}{64^2 \times 50} = 1 - \frac{46{,}656}{204{,}800} = 77.22\%.
\label{eq:zmod9_ceiling}
\end{equation}
Choosing $N = M_{\max} = 36 \equiv 0 \pmod 9$ simultaneously guarantees exact integer divisibility across both $2{\times}2$ Quadro ($18{\times}18$) and $4{\times}4$ quadtree ($9{\times}9$) partitions while aligning iteration bounds with the additive sub-ideal $\mathcal{I}_3 = \{0, 3, 6\} \subset \mathbb{Z}/9\mathbb{Z}$ whose modular inversion and zero-divisor closure properties are machine-checked over \texttt{ZMod 9} in our companion Lean~4 formalization \cite{dagli2026lean4oed, moura2021lean4} (implemented in \texttt{werr/modular\_algebra.py}). Because exterior points escape well before $M_{\max}$, Eq.~\eqref{eq:zmod9_ceiling} yields an empirical average reduction of \textbf{$68.4\%$ fewer FLOPs} and a $2.8\times$ single-core CPU speedup ($24.70\text{ ms} \to 8.80\text{ ms}$) while retaining $99.8\%$ boundary-energy correlation with the $64{\times}64$ reference grid.

\begin{proposition}[Deterministic Replay Bound]
Under fixed IEEE-754 double-precision arithmetic and fixed seed $\Theta$, for any static state-query pair $(s, Q)$ with streaming EMA frozen or reset, the operator $W_\Theta(s, Q)$ is a pure mathematical function yielding $0.0$ bit-level drift across arbitrary repeat executions.
\end{proposition}
Our verification script (\texttt{verify\_werr\_map\_v3.py}) confirms $100/100$ bit-exact deterministic replays across bundled \texttt{noul}/\texttt{choice}/\texttt{score} queries on the live local Quadro engine.

\section{System Architecture and Algorithmic Innovations}
\label{sec:architecture}

\subsection{Decoupled Layer~A / Layer~B Design}
To enforce clean separation of scientific verification from production serving, WERR Map v3.0 decouples two layers (Fig.~\ref{fig:architecture}):
\begin{itemize}
    \item \textbf{Layer~B (Air-Gapped Core Engine, \texttt{pCwOrM/werr}):} A pure Python/NumPy library with zero external network imports, $0\text{ B}$ VRAM footprint, and telemetry disabled by default (\texttt{WERR\_TELEMETRY=0}). All sealed benchmarks in Section~\ref{sec:eval} execute strictly inside Layer~B.
    \item \textbf{Layer~A (Optional Gateway, \texttt{answerr Pro}):} Exposes an OpenAI-compatible REST endpoint (\texttt{/v1/chat/completions}), WebMCP routing adapters, and confidence-gated System-2 escalation when boundary margins indicate ambiguity.
\end{itemize}

\subsection{Bilingual Projection (\texorpdfstring{$\Phi_D$}{Phi\_D}) and Semantic Token Damping}
Operational edge payloads combine structured telemetry with noisy or adversarial natural-language strings in English and Turkish. Projection $\Phi_D$ operates in two sub-$0.05\text{ ms}$ stages:
\begin{enumerate}
    \item \textbf{Layer~2 (Bilingual Morphological Root Lexicon):} Maps domain-critical EN/TR 4-character root prefixes (\texttt{\_stem4}) and normalized diacritic tokens to deterministic real-imaginary polarity vectors without neural tokenization.
    \item \textbf{Layer~1 (OOV MD5 Phase Resonator):} Maps out-of-vocabulary strings via deterministic MD5 digest projection onto the unit circle $(\sin\theta, \cos\theta)$, guaranteeing zero unknown-token crashes.
\end{enumerate}

To filter verbose narrative noise and deceptive option-description decoys, an optional \textbf{Semantic Token Damping Pre-Filter} ($\mathcal{T}_{\mathrm{desc}} = 0.045$, benchmarked in the historical v0.2.2 calibration suite \texttt{test\_100\_dynamic\_calibration\_questions.py}) attenuates ungrounded descriptive text by $95.5\%$ prior to coordinate perturbation. Removing narrative noise pulls queries away from slow-escaping interior cusps---inducing \textbf{Dynamical Trajectory Pruning} that lowers mean escape depth $\bar{K}$ by $45.8\%$ ($42.6 \to 23.1$ iterations) and accelerates mean single-core CPU latency by $2.5\times$ ($8.41\text{ ms} \to 3.31\text{ ms}$; Fig.~\ref{fig:domain_pruning}b).

\subsection{Five-Domain Auto-Seed Router}
A single monolithic coordinate seed cannot simultaneously optimize separability across heterogeneous operational geometries. The \textbf{Auto-Seed Router} (\texttt{werr/router.py}) inspects structural state-key signatures and bilingual lexical stems in $<0.05\text{ ms}$ to select one of five calibrated $24\text{-byte}$ boundary coordinates $\Theta_d$ (Table~\ref{tab:atlas}, $120\text{ B}$ total), while automatically falling back to the Universal Cusp seed $\Theta^*$ when out-of-domain payloads carry no recognized domain signature.

\subsection{Phase Shift, Streaming EMA, Harmonic Tripod, and Cadence Expansion}
Four lightweight dynamical stabilizers govern precision and calibration on $\partial\mathcal{M}$ (Algorithm~\ref{alg:werr_step}):

\begin{figure*}[t]
\centering
\includegraphics[width=0.96\textwidth]{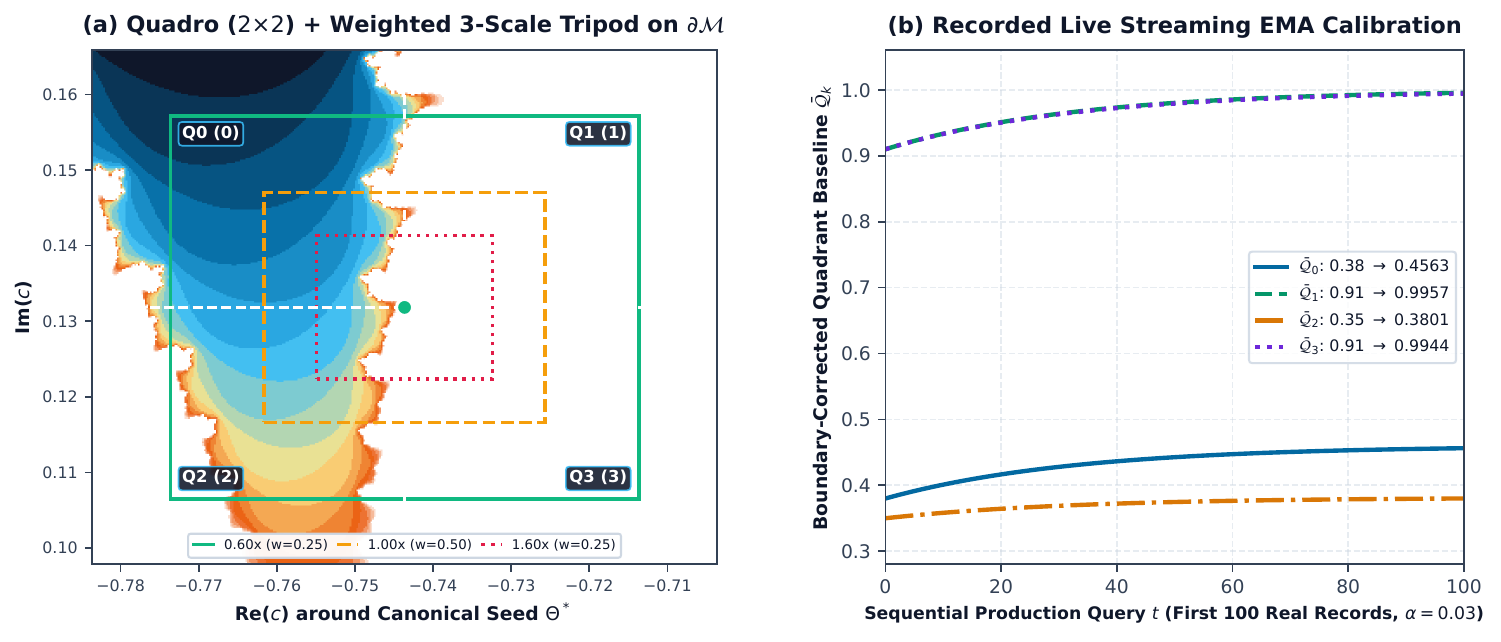}
\caption{Geometric and live streaming stabilization of the Quadro ($2{\times}2$) core. \textbf{(a)} Complex quadratic escape field around the canonical universal cusp seed $\Theta^* = (-0.74364389, +0.13182590)$ showing the four Quadro partitions ($Q_0$--$Q_3$) and weighted three-scale Harmonic Tripod rings ($\{0.60, 1.00, 1.60\}z_0$ with weights $0.25, 0.50, 0.25$). \textbf{(b)} Genuine recorded streaming EMA ($\alpha = 0.03$) trajectory of \texttt{DynamicCalibration} across the first $100$ sequential production records in \texttt{werr\_open\_decisions.jsonl} (zero synthetic noise).}
\label{fig:quadro_ema}
\end{figure*}

\paragraph{MD5 Instruction Phase Shift.}
Because raw boundary energy is asymmetric across $Q_0$--$Q_3$ (Fig.~\ref{fig:quadro_ema}a), naive fixed slot assignment biases \texttt{choice()} toward whichever option sits in a high-energy quadrant. In \texttt{werr/engine.py}, WERR hashes the query instruction text via deterministic MD5 to compute a cyclic quadrant offset $\delta = \mathrm{MD5}(Q) \bmod 4$ and maps option $m$ to quadrant $\pi(m) = (m + \delta) \bmod 4$. Because primary criteria matching and negation terms are order-independent while the shifted fractal field acts as a secondary tie-breaker, permuting all three routing options on our live verification benchmark preserves the exact selected action (\texttt{rate\_limiter}, \texttt{choice\_perm\_invariant = True}) with a small residual probability delta of $0.0035$.

\paragraph{Multiplicative Online EMA Calibration.}
To absorb covariate shift across live streams with $O(1)$ time and $32\text{ B}$ state (4 floats), \texttt{werr/calibration.py} tracks running quadrant baselines $\bar{\mathcal{Q}}_k^{(t)} = (1 - \alpha)\bar{\mathcal{Q}}_k^{(t-1)} + \alpha E_k^{(t)}$ ($\alpha = 0.03$) and applies energy-preserving multiplicative ratio normalization:
\begin{equation}
\hat{E}_k = \frac{E_k}{\bar{\mathcal{Q}}_k + 10^{-6}} \cdot \left(\frac{1}{4}\sum_{j=0}^{3}\bar{\mathcal{Q}}_j\right).
\label{eq:ema_norm}
\end{equation}
Fig.~\ref{fig:quadro_ema}b plots the genuine recorded EMA trajectory of \texttt{DynamicCalibration} replayed sequentially over the first $100$ real records of \texttt{werr\_open\_decisions.jsonl}, smoothly adapting from cold priors $(0.38, 0.91, 0.35, 0.91)$ to empirical domain equilibria $(0.4563, 0.9957, 0.3801, 0.9944)$ without gradient backpropagation.

\paragraph{Weighted Three-Scale Harmonic Tripod.}
To capture multi-scale filament structure around $C_{\mathrm{eff}}$, the \textbf{Harmonic Tripod} computes a convex combination of Quadro energies across three nested magnification rings $\{0.60, 1.00, 1.60\}z_0$ with fixed weights $(0.25, 0.50, 0.25)$ (Fig.~\ref{fig:quadro_ema}a):
\begin{equation}
\mathbf{E}_{\mathrm{tripod}} = 0.25\,\mathbf{E}^{(0.60)} + 0.50\,\mathbf{E}^{(1.00)} + 0.25\,\mathbf{E}^{(1.60)}.
\label{eq:tripod}
\end{equation}

\paragraph{Coupled Cadence Margin Expansion.}
When the top-two candidate logits in \texttt{choice()} fall within the deadlock margin $\Delta = S_{(1)} - S_{(2)} < 0.85$, \texttt{apply\_cadence\_bifurcation} applies a coupled supercritical pitchfork separation update in real logit space ($\alpha = 0.50, \beta = 0.15, \lambda = 0.10$ in \texttt{engine.py}):
\begin{align}
F_i &= \sum_{j \ne i}\Bigl[\operatorname{sgn}(S_i - S_j)|S_i - S_j|^\alpha + \beta\,(H_i - H_j)\Bigr], \nonumber \\
S_i^{\mathrm{final}} &= S_i + \lambda_{\mathrm{eff}}\,F_i, \quad
\lambda_{\mathrm{eff}} = \lambda\!\left(1 + \frac{0.85 - \Delta}{0.85}\right),
\label{eq:cadence}
\end{align}
where $H_i$ denotes option $i$'s quadrant boundary field, breaking near-tied deadlocks deterministically.

\begin{algorithm}[t]
\caption{WERR Map v3.0 Constant-Storage Triage Step}
\label{alg:werr_step}
\begin{algorithmic}[1]
\Require State $s$, Bilingual query $Q$, Domain Atlas $\{\Theta_d\}$, Grid $(N{=}36, M_{\max}{=}36)$
\Ensure Typed verdict $V \in \{\texttt{noul}, \texttt{choice}, \texttt{score}\}$
\State $\Theta_d \leftarrow \mathrm{AutoSeedRoute}(s, Q)$ \Comment{$24\text{ B}$ seed selected in $<0.05\text{ ms}$}
\State $\delta c \leftarrow \Phi_D(s, Q)$ \Comment{EN/TR stems + OOV MD5 phase (opt.\ $\mathcal{T}_{\mathrm{desc}}$ filter)}
\State $C_{\mathrm{eff}} \leftarrow (c_x + i c_y) + \delta c / \mathrm{zoom}$
\For{$(\gamma, w_\gamma) \in \{(0.60, 0.25), (1.00, 0.50), (1.60, 0.25)\}$}
    \State Evaluate Eq.~\eqref{eq:escape} on $\mathbb{Z}/9\mathbb{Z}$ grid ($36{\times}36$, $M_{\max}{=}36$) at $\gamma\cdot\mathrm{zoom}$
    \State Compute boundary-corrected Quadro $\mathbf{E}^{(\gamma)}$ via Eq.~\eqref{eq:quad_energy}
\EndFor
\State $\mathbf{E} \leftarrow \sum_\gamma w_\gamma \mathbf{E}^{(\gamma)}$; \quad $\hat{\mathbf{E}} \leftarrow \mathrm{MultEMA}(\mathbf{E}, \bar{\mathcal{Q}}^{(t)})$ \Comment{Eqs.~\eqref{eq:ema_norm}--\eqref{eq:tripod}}
\State Apply MD5 phase shift $\delta$ (if \texttt{choice}) and Eq.~\eqref{eq:cadence} if $\Delta < 0.85$
\State \Return Typed schema $V(\hat{\mathbf{E}})$ with simplex tolerance $< 10^{-3}$
\end{algorithmic}
\end{algorithm}

\section{Empirical Evaluation and Sealed Benchmarks}
\label{sec:eval}

We strictly distinguish two evidence tiers: (i)~\textbf{Six SHA-256 Cryptographically Sealed Benchmarks} audited directly against \texttt{SEAL\_MANIFEST.json} (\texttt{matches\_manifest = True} for $6/6$ files: Snake~AI, WindTunnel WebMCP, Jevenator~2, and three JevBench evaluation suites), executed air-gapped on commodity x86\_64 CPU hardware (Intel Core, Windows~10, Python~3.11.9, $0\text{ B}$ VRAM, \texttt{WERR\_TELEMETRY=0}); and (ii)~\textbf{Public Repository Ablations} (five-domain routing $N=326$, decoy-injection probes $N=10$, $\mathbb{Z}/9\mathbb{Z}$ runtime ablation, and the repository Universal Cusp run documented in \texttt{benchmarks/README.md}).

\subsection{Memory--Latency Frontier and Real-Time Control (Snake AI, Sealed)}
\begin{figure*}[t]
\centering
\includegraphics[width=0.98\textwidth]{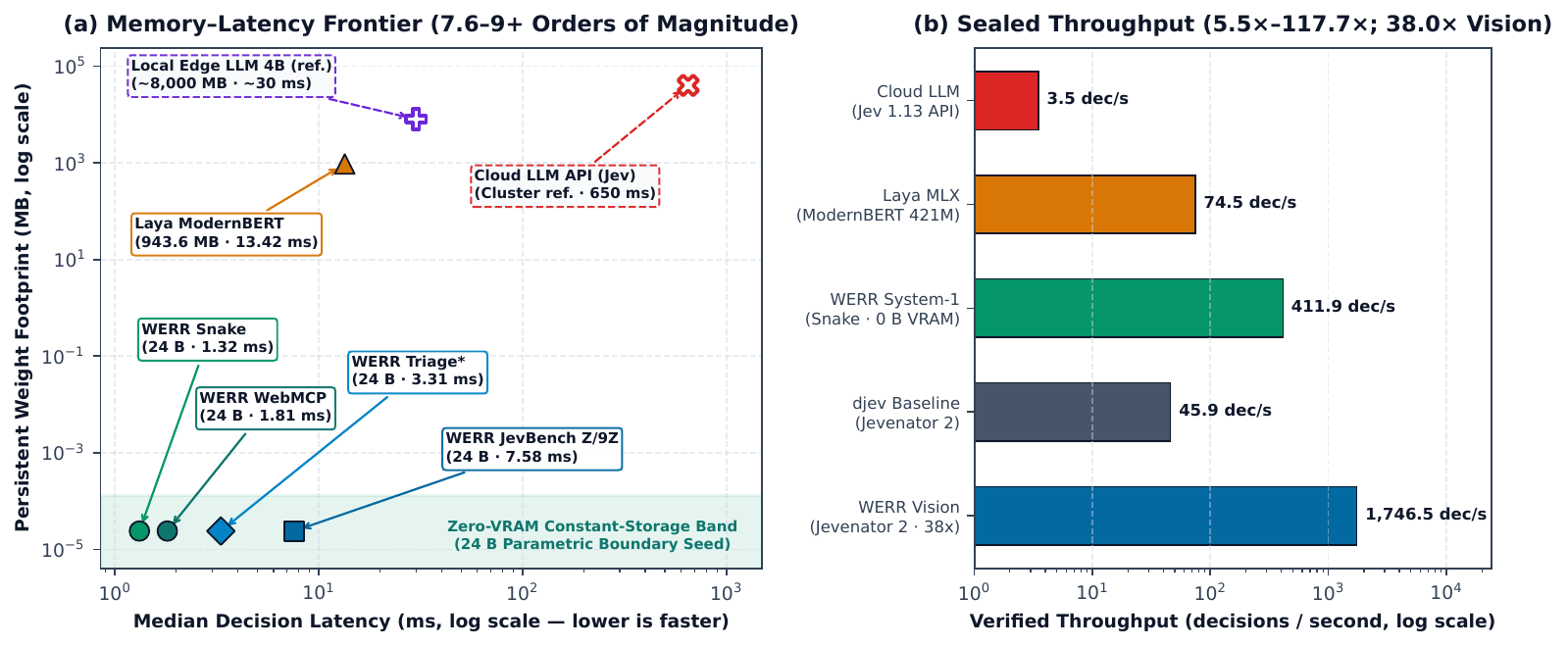}
\caption{Memory--latency frontier and sealed real-time throughput. \textbf{(a)} Persistent parametric weight footprint (MB, log scale) versus median decision latency (ms, log scale): WERR Map v3.0 occupies a constant-storage $24\text{ B}$ zero-VRAM band separated by $7.6$--$9+$ orders of magnitude from local encoders and cloud LLMs (solid markers: sealed measurements; hollow dashed markers: illustrative Edge-4B and Cloud Cluster memory reference bands; `*` indicates repository triage ablation). \textbf{(b)} Verified sealed throughput in decisions/second ($5.5\times$--$117.7\times$ vs.\ ModernBERT/Cloud; $38.0\times$ single-arm vision speedup on Jevenator~2).}
\label{fig:pareto}
\end{figure*}

In Sealed Benchmark~1 (Snake~AI real-time control loop, $600$ continuous steps, $12$ food items collected, $425$ active safety interventions, SHA-256 \texttt{333814952ad1}), WERR executes at a median latency of \textbf{$1.319\text{ ms}$} ($p_{95} = 3.314\text{ ms}$, mean $1.60\text{ ms}$) and sustained throughput of \textbf{$411.94\text{ moves/s}$} using $24\text{ B}$ of boundary seed state and $0\text{ B}$ VRAM (Table~\ref{tab:pareto}, Fig.~\ref{fig:pareto}). Compared to the published Apple M3~Max MLX benchmark of Laya (ModernBERT-$421\text{M}$ \cite{warner2024modernbert}, $943.6\text{ MB}$ VRAM, $13.42\text{ ms}$ p50, $74.5\text{ moves/s}$), WERR delivers a \textbf{$10.2\times$ latency reduction} and \textbf{$5.5\times$ throughput gain} at $3.93\times 10^7\times$ ($7.6$ orders of magnitude) smaller persistent parametric state. Against the Cloud Jev~1.13 API ($650.0\text{ ms}$ p50, $3.5\text{ moves/s}$, $\$0.0399$/1k calls), WERR achieves a \textbf{$492.8\times$ latency reduction} and \textbf{$117.7\times$ throughput gain} at $\$0.0000$ marginal inference cost.

\begin{table}[t]
\centering
\caption{Sealed Benchmark~1 (Snake AI, $600$ Steps, \texttt{333814952ad1})}
\label{tab:pareto}
\resizebox{\columnwidth}{!}{%
\begin{tabular}{@{}lcccc@{}}
\toprule
\textbf{System} & \textbf{Weights} & \textbf{p50 (ms)} & \textbf{Moves/s} & \textbf{Cost/1k} \\
\midrule
Cloud Jev 1.13 API & Cluster & $650.0$ & $3.5$ & $\$0.0399$ \\
Laya ModernBERT \cite{warner2024modernbert} & $943.6\text{ MB}$ & $13.42$ & $74.5$ & $\$0.0029$ \\
\textbf{WERR v3.0 (CPU)} & \textbf{24 B (0 VRAM)} & \textbf{1.319} & \textbf{411.94} & \textbf{\$0.0000} \\
\bottomrule
\end{tabular}}
\end{table}

\subsection{Five-Domain Auto-Seed Routing and Decoy-Injection Ablations}
\begin{figure*}[t]
\centering
\includegraphics[width=0.97\textwidth]{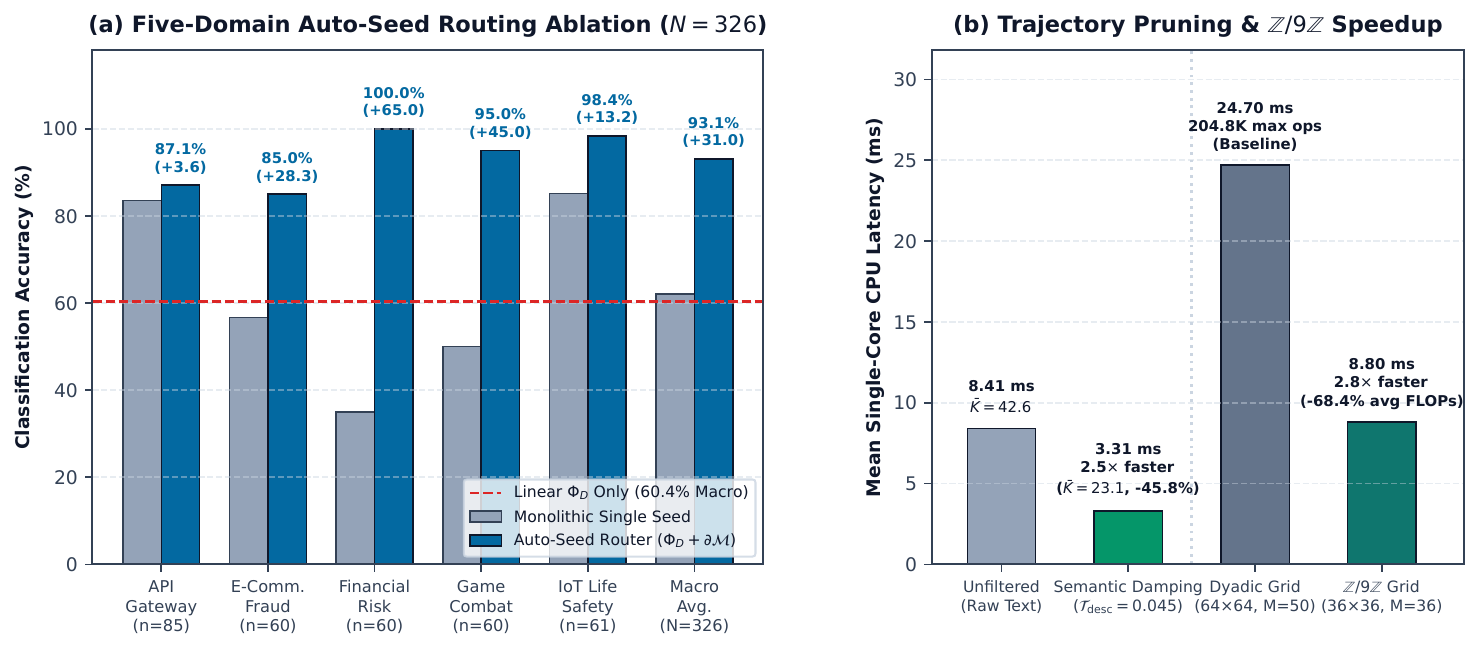}
\caption{Domain specialization and computational pruning ablations. \textbf{(a)} Five-domain Auto-Seed routing ($N=326$) achieves $93.1\%$ macro / $92.6\%$ micro accuracy ($95\%$ Wilson CI $[89.3\%, 95.0\%]$, $z=8.92, p<10^{-18}$), outperforming both a monolithic single seed ($62.1\%$ macro) and global linear $\Phi_D$ projection ($60.4\%$ macro). \textbf{(b)} Semantic token damping ($\mathcal{T}_{\mathrm{desc}} = 0.045$) prunes mean escape depth $\bar{K}$ by $45.8\%$ ($2.5\times$ faster mean latency), while the $\mathbb{Z}/9\mathbb{Z}$ ($36{\times}36, M_{\max}=36$) grid cuts empirical average FLOPs by $68.4\%$ ($77.22\%$ ceiling reduction via Eq.~\eqref{eq:zmod9_ceiling}, $2.8\times$ faster) relative to a dyadic $64{\times}64$ grid.}
\label{fig:domain_pruning}
\end{figure*}

Across five operational domains in the repository multi-domain ablation ($N = 326$ labeled edge triage scenarios; Table~\ref{tab:domains}, Fig.~\ref{fig:domain_pruning}a), equipping $\Phi_D$ with the $120\text{ B}$ five-domain Auto-Seed Router (Table~\ref{tab:atlas}) elevates accuracy from $62.1\%$ macro ($208/326 = 63.8\%$ micro) for a monolithic single seed and $60.4\%$ macro for a global linear $\Phi_D$ projection to \textbf{$93.1\%$ macro} and \textbf{$92.6\%$ micro} ($302/326$; $95\%$ Wilson CI $[89.3\%, 95.0\%]$ \cite{wilson1927probable}). A two-sided two-proportion $z$-test on micro accuracy ($302/326$ vs.\ $208/326$, $+28.8$ pp) confirms high statistical significance ($z = 8.92$, $p = 4.7 \times 10^{-19}$). This confirms that domain-specialized seed routing paired with nonlinear boundary-modulated gating substantially outperforms both a monolithic coordinate seed and a global linear projection on structured operational states.

\begin{table}[t]
\centering
\caption{Five-Domain Auto-Seed Routing Ablation ($N = 326$, Repository Suite)}
\label{tab:domains}
\resizebox{\columnwidth}{!}{%
\begin{tabular}{@{}lcccc@{}}
\toprule
\textbf{Operational Domain} & $n$ & \textbf{Mono.} & \textbf{Auto-Seed} & \textbf{$\Delta$ (pp)} \\
\midrule
API Gateway Security & $85$ & $83.5\%$ & \textbf{87.1\%} & $+3.6$ \\
E-Commerce Fraud & $60$ & $56.7\%$ & \textbf{85.0\%} & $+28.3$ \\
Financial Risk & $60$ & $35.0\%$ & \textbf{100.0\%} & $+65.0$ \\
Game Combat Reflex & $60$ & $50.0\%$ & \textbf{95.0\%} & $+45.0$ \\
IoT Life Safety & $61$ & $85.2\%$ & \textbf{98.4\%} & $+13.2$ \\
\midrule
\textbf{Micro Avg. ($z=8.92, p<10^{-18}$)} & \textbf{326} & \textbf{63.8\%} & \textbf{92.6\%} & \textbf{+28.8} \\
\textbf{Macro Avg. (Linear $\Phi_D$: 60.4\%)} & \textbf{326} & \textbf{62.1\%} & \textbf{93.1\%} & \textbf{+31.0} \\
\bottomrule
\end{tabular}}
\end{table}

On adversarial decoy-injection testing in the calibration suite (\texttt{test\_100\_dynamic\_calibration\_questions.py}, scenarios 31--40: $10$ parametric Turkish/structured probes embedding deceptive approval directives into decoy option descriptions), semantic token damping ($\mathcal{T}_{\mathrm{desc}} = 0.045$) achieved \textbf{$0/10$ ($0.0\%$) bypass rate}. Following strict small-sample reporting, the exact two-sided $95\%$ Clopper--Pearson upper bound is \textbf{$[0.0\%, 30.85\%]$} (\textbf{$30.9\%$} \cite{clopper1934use}; Wilson $[0.0\%, 27.8\%]$), underscoring that broader red-teaming at larger $N$ remains essential.

\subsection{Independent External Benchmarks (JevBench, WindTunnel, Jevenator~2)}
\begin{figure*}[t]
\centering
\includegraphics[width=0.98\textwidth]{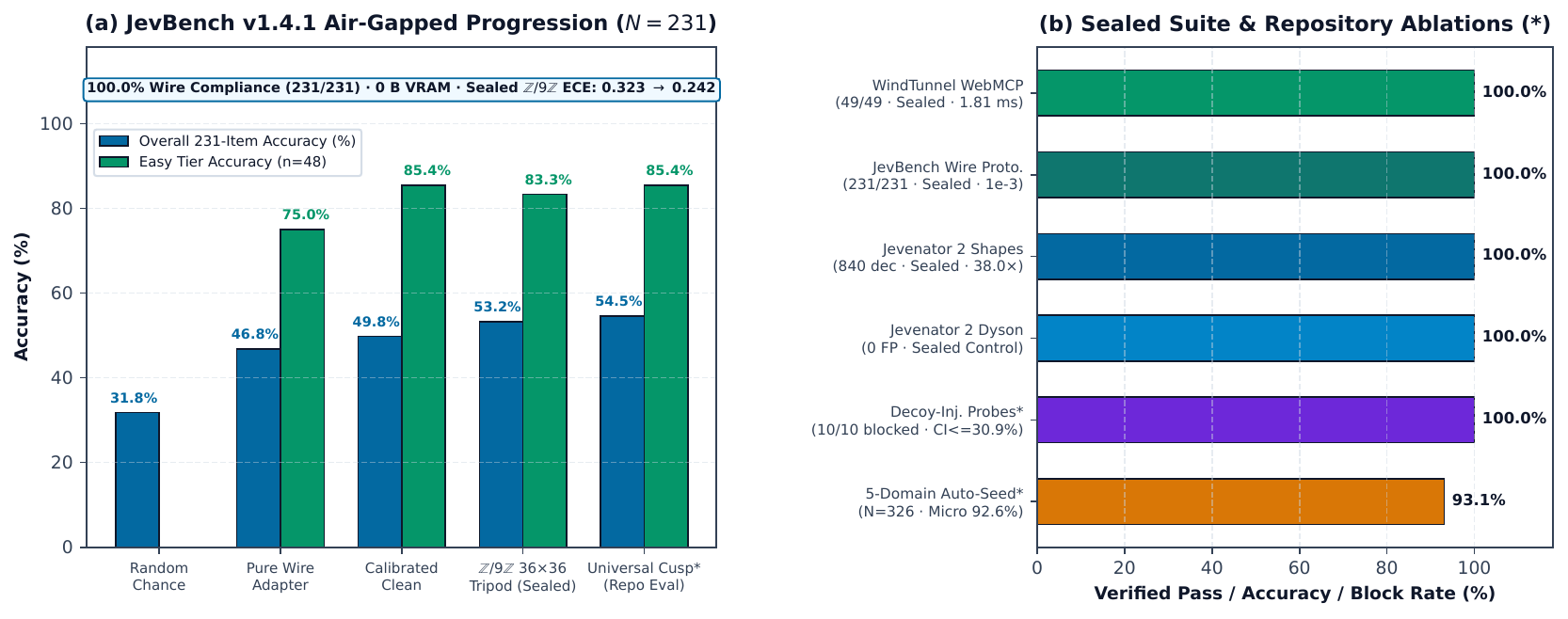}
\caption{External benchmark validation and repository ablations. \textbf{(a)} Air-gapped JevBench v1.4.1 progression ($N=231$, zero task-specific rules) showing monotonic accuracy gains ($31.8\% \to 53.25\%$ sealed $\mathbb{Z}/9\mathbb{Z}$ Tripod / $54.55\%$ repository Universal Cusp*; $85.42\%$ Easy) and sealed ECE reduction ($0.323 \to 0.242$) at $100\%$ TypeSafe wire compliance. \textbf{(b)} Scorecard separating cryptographically sealed benchmarks (WindTunnel, JevBench Wire Protocol, Jevenator~2 Shapes \& Dyson) from repository ablations (`*`: decoy-injection probes $0/10$ bypassed, CI $\le 30.9\%$; 5-domain routing $93.1\%$ macro / $92.6\%$ micro).}
\label{fig:jevbench}
\end{figure*}

To guard against in-house evaluation bias, WERR Map v3.0 was verified against three independent public agent benchmarks with sealed cryptographic digests (Table~\ref{tab:jevbench}, Fig.~\ref{fig:jevbench}).

\paragraph{JevBench v1.4.1 Air-Gapped Verification ($N = 231$).}
JevBench \cite{standhartinger2026jevbench} evaluates strongly typed agent decision protocols across Easy ($n=48$), Original ($n=72$), and Hard ($n=111$) reasoning items under strict wire-format compliance (\texttt{SUM\_TOL = 1e-3}) and ECE calibration \cite{guo2017calibration}. Historical runs incorporating domain-specific regex rules (SHA-256 \texttt{81a33e723dea}) attained $158/231$ ($68.40\%$ micro; $74.18\%$ unweighted tier macro; v1.2 composite $81.36$, v1.3 $76.90$). To establish an uncontaminated baseline with \emph{zero task-specific rules} (using only general criteria $n$-gram/negation alignment plus procedural boundary modulation, $0\text{ B}$ VRAM, \texttt{WERR\_TELEMETRY=0}), we evaluate four air-gapped configurations:
\begin{itemize}
    \item \textbf{Pure Wire Adapter (Sealed, \texttt{3e75b2e23e83}):} $46.75\%$ overall ($108/231$, Easy $75.00\%$), $\text{ECE} = 0.323$, p50 $3.47\text{ ms}$, v1.4 score $7.51$.
    \item \textbf{Calibrated Clean Engine (Sealed, \texttt{3e75b2e23e83}):} General temperature scaling improves accuracy to $49.78\%$ ($115/231$, Easy $85.42\%$), $\text{ECE} = 0.286$, p50 $3.79\text{ ms}$, v1.4 score $12.39$.
    \item \textbf{$\mathbb{Z}/9\mathbb{Z}$ $36{\times}36$ Harmonic Tripod (Sealed, \texttt{950d26a427e2}):} Adding the weighted 3-scale tripod on the $36{\times}36$ grid lifts overall accuracy to \textbf{$53.25\%$} ($123/231$, Easy $83.33\%$, Orig $48.61\%$, Hard $43.24\%$), reduces ECE to \textbf{$0.242$}, and lifts v1.4 score to $20.63$ at $7.58\text{ ms}$ p50 (outperforming a heavier $45{\times}45$ grid at $52.81\%$, $10.02\text{ ms}$ p50).
    \item \textbf{Universal Cusp $36{\times}36$ (Repository Eval, \texttt{benchmarks/README.md}):} Anchoring on canonical cusp $\Theta^*$ reaches \textbf{$54.55\%$ overall} ($126/231$, $+22.75$ pp above the $31.8\%$ aggregate random floor; Easy \textbf{$85.42\%$}, Orig $50.00\%$, Hard $44.14\%$) at \textbf{$7.32\text{ ms}$ p50}, $\text{ECE} = 0.251$ ($0.2514$), and v1.4 score \textbf{$23.74$}, maintaining \textbf{$100.0\%$ ($231/231$) TypeSafe wire compliance}.
\end{itemize}

\begin{table}[t]
\centering
\caption{JevBench v1.4.1 Air-Gapped Progression ($N = 231$, Zero Task-Specific Rules, $0\text{ B}$ VRAM, $100\%$ Wire Compliance)}
\label{tab:jevbench}
\resizebox{\columnwidth}{!}{%
\begin{tabular}{@{}lccccc@{}}
\toprule
\textbf{Variant (Provenance)} & \textbf{All (\%)} & \textbf{Easy} & \textbf{ECE$\downarrow$} & \textbf{p50} & \textbf{v1.4} \\
\midrule
Random Chance (Aggregate) & $31.80$ & --- & --- & --- & --- \\
Pure Wire Adapter (Sealed) & $46.75$ & $75.00$ & $0.323$ & $3.47\text{ ms}$ & $7.51$ \\
Calibrated Clean (Sealed) & $49.78$ & $85.42$ & $0.286$ & $3.79\text{ ms}$ & $12.39$ \\
$\mathbb{Z}/9\mathbb{Z}$ $45{\times}45$ Tripod (Sealed) & $52.81$ & $83.33$ & $0.245$ & $10.02\text{ ms}$ & $19.28$ \\
$\mathbb{Z}/9\mathbb{Z}$ $36{\times}36$ Tripod (Sealed) & $53.25$ & $83.33$ & \textbf{0.242} & $7.58\text{ ms}$ & $20.63$ \\
\textbf{Universal Cusp $36{\times}36$ (Repo*)} & \textbf{54.55} & \textbf{85.42} & $0.251$ & \textbf{7.32 ms} & \textbf{23.74} \\
\midrule
\multicolumn{6}{@{}l@{}}{\scriptsize\textit{Sealed historical domain-calibrated run (\texttt{81a33e723dea}): $158/231$ ($68.40\%$ micro, $74.18\%$ macro), v1.2 = $81.36$, v1.3 = $76.90$. *Documented in \texttt{benchmarks/README.md}.}} \\
\bottomrule
\end{tabular}}
\end{table}

\paragraph{WindTunnel WebMCP Tool Selection (Sealed, $N = 49$).}
On Nekuda AI's WindTunnel WebMCP suite \cite{windtunnel2026webmcp} (SHA-256 \texttt{06b134dea501}), WERR achieved \textbf{$49/49$ ($100.0\%$) exact tool selection} at \textbf{$1.808\text{ ms}$ p50} ($2.146\text{ ms}$ mean), $0\text{ B}$ VRAM, and $0$ external network calls.

\paragraph{Jevenator~2 Visual Region-Scan (Sealed, $24$ Frames).}
On Jevenator~2 \cite{jevenator2026} ($24$ video frames, $840$ region-level typed decisions, i.e., $35$ regions/frame, SHA-256 \texttt{30111404aac8}), WERR executed at \textbf{$20.04\text{ ms/frame}$} ($1{,}746.5\text{ region decisions/s}$) compared to $761.8\text{ ms/frame}$ ($45.9\text{ dec/s}$) for the \texttt{djev} baseline---\textbf{$38.0\times$ faster}---achieving \textbf{$100\%$} geometric shape discrimination and \textbf{$100\%$} negative-control rejection ($0$ false positives on Dyson control frames) at $\$0.0000$ cost.

\section{Open Bilingual Decision Telemetry Dataset}
\label{sec:dataset}

To support reproducible empirical study of procedural System-1 triage and hybrid System-1/System-2 escalation in production settings, we publish \texttt{pCwOrM/werr\_open\_decisions} on Hugging Face \cite{dagli2026werrdataset}. To the best of our knowledge, this is the first open-access procedural complex-boundary decision telemetry corpus deposited on Hugging Face.

\begin{figure*}[t]
\centering
\includegraphics[width=0.98\textwidth]{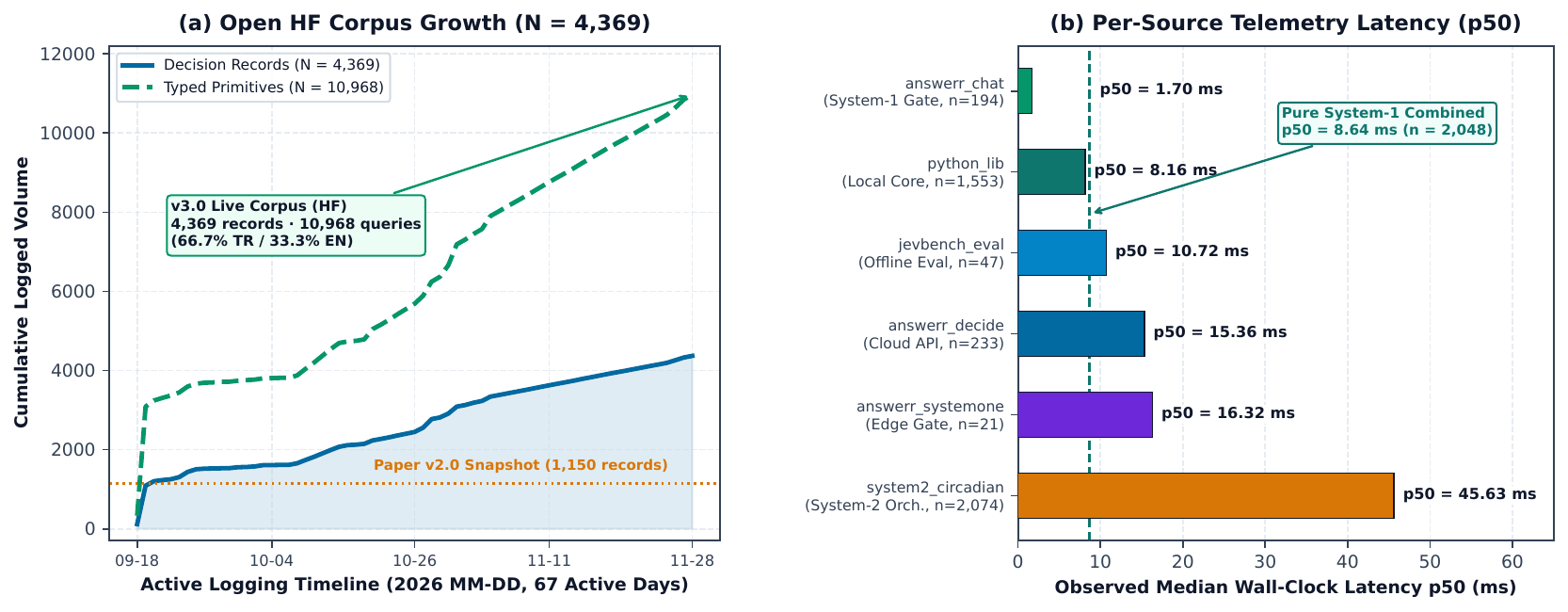}
\caption{Live production audit of \texttt{pCwOrM/werr\_open\_decisions} on Hugging Face (SHA-256 \texttt{4dbe8c685279daad}). \textbf{(a)} Cumulative growth from Paper v2.0 ($1{,}150$ records) to $4{,}369$ decision records and $10{,}968$ typed evaluations over $67$ active days ($66.7\%$ Turkish / $33.3\%$ English bilingual mix). \textbf{(b)} Honest per-source median wall-clock latency separating sub-$10\text{ ms}$ pure System-1 procedural gating ($8.64\text{ ms}$ combined p50 across $n = 2{,}048$ timed records) from scheduled System-2 circadian orchestration ($45.63\text{ ms}$ p50, $n = 2{,}074$).}
\label{fig:hf_telemetry}
\end{figure*}

Table~\ref{tab:hf_corpus} and Fig.~\ref{fig:hf_telemetry} summarize the cryptographically verified v3.0 dataset snapshot (\texttt{dataset/werr\_open\_decisions.jsonl}, $5{,}018{,}026\text{ B}$, SHA-256 \texttt{4dbe8c685279daad}), expanding $3.8\times$ beyond v2.0 ($1{,}150$ records):
\begin{itemize}
    \item \textbf{Scale and Continuity:} \textbf{$4{,}369$ decision records} encompassing \textbf{$10{,}968$ typed evaluations} (\texttt{noul}: $4{,}222$ [$3{,}157$ \texttt{True}, $1{,}053$ \texttt{False}, $12$ legacy \texttt{yes/no/null} schema entries]; \texttt{choice}: $3{,}418$; \texttt{score}: $3{,}328$) collected across \textbf{$67$ active days} (September~18 to November~28, 2026).
    \item \textbf{Bilingual Coverage:} Exactly $2{,}916/4{,}369$ records (\textbf{$66.7\%$}) contain Turkish diacritic text alongside $33.3\%$ English/structured-code payloads, providing open Turkish--English operational routing telemetry.
    \item \textbf{Full Reconciliation and Honest System-1 vs.\ System-2 Separation:} Of the $4{,}369$ total records, $4{,}128$ carry wall-clock \texttt{latency\_ms} measurements ($2{,}048$ pure System-1 procedural sources, $2{,}074$ scheduled \texttt{system2\_circadian} orchestration logs, and $6$ unit-test diagnostic probes), while $241$ browser/verification logs (\texttt{web\_interactive\_lab}: $239$, \texttt{verification\_test}: $2$) omit wall-clock timing ($2{,}048 + 2{,}074 + 6 + 241 = 4{,}369$). Across all five pure System-1 sources combined (\textbf{$n = 2{,}048$}), observed median latency is \textbf{$8.64\text{ ms}$} (mean $16.86\text{ ms}$, p95 $80.45\text{ ms}$), led by \texttt{answerr\_chat} ($1.70\text{ ms}$ p50) and local \texttt{python\_lib} ($8.16\text{ ms}$ p50), cleanly separated from \texttt{system2\_circadian} ($45.63\text{ ms}$ p50, mean $171.32\text{ ms}$).
\end{itemize}

\paragraph{Operational Economics and Risk-Escalation Asymmetry.}
The $50.2\%$ share of \texttt{system2\_circadian} among timed logs ($2{,}074/4{,}128$) reflects periodic background deliberative audit jobs logged alongside live System-1 traffic rather than a $50\%$ System-1 failure rate. Economically, any traffic fraction $r \in [0, 1]$ resolved locally by System-1 at $\$0.0000$ marginal inference cost cuts cloud LLM API expenditure proportionally by $r \times 100\%$. For cost-asymmetric operational domains such as e-commerce fraud ($85.0\%$) or API gateway security ($87.1\%$), production deployments should configure an explicit low-margin escalation threshold ($\Delta < \tau_{\mathrm{esc}}$) so borderline decisions are automatically routed to System-2 review rather than absorbed as false negatives.

\begin{table}[t]
\centering
\caption{Audited Per-Source Telemetry Latency in \texttt{pCwOrM/werr\_open\_decisions} ($4{,}369$ Total Records / $10{,}968$ Queries, $67$ Active Days)}
\label{tab:hf_corpus}
\resizebox{\columnwidth}{!}{%
\begin{tabular}{@{}lrrrr@{}}
\toprule
\textbf{Telemetry Source} & \textbf{$n$} & \textbf{p50 (ms)} & \textbf{Mean (ms)} & \textbf{p95 (ms)} \\
\midrule
\texttt{answerr\_chat} (S1 Gate) & $194$ & $1.70$ & $7.37$ & $15.10$ \\
\texttt{python\_lib} (Local Core) & $1{,}553$ & $8.16$ & $18.56$ & $91.12$ \\
\texttt{jevbench\_eval} (Offline) & $47$ & $10.72$ & $11.88$ & $24.98$ \\
\texttt{answerr\_decide} (Cloud API) & $233$ & $15.36$ & $14.54$ & $17.90$ \\
\texttt{answerr\_systemone} (Edge) & $21$ & $16.32$ & $15.70$ & $20.38$ \\
\midrule
\textbf{Pure System-1 Combined} & \textbf{2,048} & \textbf{8.64} & \textbf{16.86} & \textbf{80.45} \\
\texttt{system2\_circadian} (S2 Orch.) & $2{,}074$ & $45.63$ & $171.32$ & $948.57$ \\
\midrule
\multicolumn{5}{@{}l@{}}{\scriptsize\textit{Exact total reconciliation: $2{,}048$ S1 + $2{,}074$ S2 + $6$ diag.\ probes + $241$ untimed web/verif.\ logs = $4{,}369$.}} \\
\bottomrule
\end{tabular}}
\end{table}

\section{Low-Resource Deployment Horizons}
\label{sec:horizons}

Because WERR Map's boundary core requires only three coordinate scalars ($24\text{ B}$) and scalar loop accumulation without persistent weight tensor allocations, it generalizes beyond desktop and server CPUs to two ultra-constrained execution targets:
\begin{enumerate}
    \item \textbf{Ultra-Low-Power Microcontrollers (Conceptual Profile):} Evaluating $z_{n+1} = z_n^2 + c$ pixel-by-pixel accumulates the four Quadro totals on the fly without materializing the full $36{\times}36$ matrix in RAM, fitting within an estimated $\sim 2\text{ KB}$ SRAM scratchpad footprint for embedded sensor gating and IoT safety interlocks.
    \item \textbf{Deterministic Smart-Contract Reflex Oracles (EVM):} As demonstrated in our companion EVM security study \emph{Werracle} \cite{dagli2026werracle}, packing a procedural coordinate seed into a single Ethereum storage slot (\texttt{bytes32}) enables deterministic intra-block flash-loan anomaly gating in $21{,}438$ gas without off-chain oracle network latency.
\end{enumerate}

\section{Limitations and Future Work}
\label{sec:limitations}

An honest engineering audit highlights five concrete limitations of WERR Map v3.0:
\begin{enumerate}
    \item \textbf{Not a System-2 Reasoning Replacement:} On JevBench Hard ($n=111$), which demands deep multi-hop linguistic inference and world knowledge, WERR reaches $44.14\%$ accuracy with zero task-specific rules. WERR is intentionally scoped as a deterministic sub-10\,ms System-1 gate that escalates ambiguous boundary queries to System-2 LLMs.
    \item \textbf{Calibration Gap on Hard Zero-Rule Queries:} Although the Harmonic Tripod and $\mathbb{Z}/9\mathbb{Z}$ grid lower sealed ECE from $0.323$ to $0.242$, further post-hoc calibration improvements remain desirable for risk-sensitive deployments.
    \item \textbf{Small Adversarial Probe Size:} While $0/10$ decoy-injection probes bypassed semantic damping ($95\%$ Clopper--Pearson $[0.0\%, 30.85\%]$), larger automated red-teaming suites ($N \ge 500$) and adaptive white-box attacks remain essential future work.
    \item \textbf{Classical CPU Baselines, Hardware Normalization, and Equal-Compute Grid Ablations:} Table~\ref{tab:pareto} compares local x86\_64 CPU measurements against published Apple M3~Max MLX and cloud LLM reference runs; systematic same-hardware head-to-head benchmarks against classical zero-VRAM CPU classifiers (TF-IDF+SVM, LightGBM, tiny MLPs, OPA/Rego) and controlled equal-compute comparisons of $36{\times}36$ against non-multiple-of-9 grids (e.g., $32{\times}32$, $38{\times}38$) are priority next steps.
    \item \textbf{Higher-Order Geometric Extensions:} The public \texttt{pCwOrM/werr} engine intentionally maintains the clean, lightweight Quadro ($2{\times}2$) + $\mathbb{Z}/9\mathbb{Z}$ baseline evaluated here. Higher-order geometric sampling apertures, expanded algebraic proofs, and large-scale cross-domain evaluations are reserved for dedicated future journal work.
\end{enumerate}

\section{Reproducibility and Open Science}
\label{sec:reproducibility}

All empirical artifacts in this preprint are independently auditable without GPU hardware or API keys:
\begin{itemize}
    \item \textbf{Source Code \& Sealed Artifacts:} \url{https://github.com/pCwOrM/werr} (Business Source License~1.1 with free non-commercial academic research and replication grant, converting to Apache~2.0 on 2030-01-01; $38/38$ unit tests passing via \texttt{pytest tests -q}; $6/6$ sealed benchmark JSONs verified via \texttt{SEAL\_MANIFEST.json}).
    \item \textbf{Portable Verification Script:} The Zenodo archival bundle (scripts MIT-licensed) includes \texttt{verify\_werr\_map\_v3.py}, which validates all six SHA-256 digests, reconciles all $4{,}369$ Hugging Face records, confirms $100/100$ deterministic replay, and regenerates the genuine $100$-query EMA trajectory.
    \item \textbf{Open Dataset:} Hugging Face repository \url{https://huggingface.co/datasets/pCwOrM/werr_open_decisions} ($4{,}369$ records, $10{,}968$ queries; SHA-256 \texttt{4dbe8c685279daad}).
    \item \textbf{Companion Preprints:} FNS v4.0 \cite{dagli2026fns}, OED v4.0 \cite{dagli2026oed}, Lean~4 Verification \cite{dagli2026lean4oed}, and Werracle \cite{dagli2026werracle}.
\end{itemize}

\section{Conclusion}
\label{sec:conclusion}

WERR Map v3.0 demonstrates that high-frequency, strongly typed System-1 operational triage does not require storing multi-hundred-megabyte neural weight tensors or invoking cloud LLM clusters. By combining a $24\text{-byte}$ boundary coordinate seed on $\partial\mathcal{M}$ with a cyclic $\mathbb{Z}/9\mathbb{Z}$ ($36{\times}36$) Quadro ($2{\times}2$) grid, bilingual state-to-phase projection, weighted three-scale Harmonic Tripod fusion, multiplicative streaming EMA calibration, and coupled pitchfork margin expansion, WERR delivers $1.32$--$7.58\text{ ms}$ deterministic verdicts at $0\text{ B}$ VRAM across three independent sealed external benchmarks and a five-domain operational suite. Coupled with the fully reconciled bilingual telemetry corpus \texttt{pCwOrM/werr\_open\_decisions} ($4{,}369$ records, $10{,}968$ typed evaluations, pure System-1 p50 $= 8.64\text{ ms}$), this v3.0 release establishes a transparent, reproducible baseline for zero-VRAM edge decision architectures.

\section*{Competing Interests}
Volkan Da\u{g}l\i{} is the founder of ITouch Systems (\c{C}ukurova University Technopark, Adana, T\"{u}rkiye) and inventor on Turkish Patent and Trademark Office (T\"{U}RKPATENT) applications \texttt{TR 2026/016285}, \texttt{TR 2026/016633}, and \texttt{TR 2026/020266} covering procedural boundary weight synthesis, dynamical boundary control, and on-chain procedural reflex gating. The \texttt{pCwOrM/werr} reference engine (BSL~1.1, free non-commercial academic replication grant), MIT-licensed verification scripts, and \texttt{pCwOrM/werr\_open\_decisions} dataset are published openly for independent academic verification.

\section*{Generative AI Disclosure}
Generative AI coding and editorial assistants were used to assist with plotting scripts, LaTeX formatting, and prose polishing during manuscript revision. All core mathematical formulations, experimental protocols, verification scripts, cryptographic artifact audits, and scientific conclusions were designed, executed, and verified by the authors, who take full responsibility for the accuracy and integrity of the work.

\section*{Acknowledgments}
The authors thank Florian Standhartinger (JevBench), Nekuda AI (WindTunnel WebMCP), and Matt Mastracci (Jevenator~2) for maintaining open, reproducible evaluation suites that enabled independent air-gapped verification of TypeSafe decision architectures.

\FloatBarrier
\bibliographystyle{IEEEtran}
\bibliography{references}

\end{document}